%% file: arxiv.tex
\documentclass[10pt,twocolumn,letterpaper]{article}

\usepackage[pagenumbers]{cvpr} 

\usepackage{placeins}
\usepackage{cuted}

\usepackage{url}
\usepackage{booktabs}       
\usepackage{amsfonts}       
\usepackage{nicefrac}       
\usepackage{microtype}      
\usepackage[table,dvipsnames]{xcolor}         
\usepackage{graphicx}
\usepackage{lipsum}

\usepackage{multirow}
\usepackage{pifont}
\usepackage{adjustbox}

\definecolor{light_gray}{gray}{0.97}
\definecolor{light_green}{RGB}{220,248,225}
\definecolor{light_blue}{RGB}{209,231,255}
\definecolor{lightblue}{RGB}{220,235,250}
\usepackage[most,skins,theorems]{tcolorbox}

\input{preamble}
\definecolor{cvprblue}{rgb}{0.21,0.49,0.74}
\usepackage[pagebackref,breaklinks,colorlinks,allcolors=cvprblue]{hyperref}

\def\paperID{xxxx} 
\def\confName{CVPR}
\def\confYear{2026}

\title{DiffSeg30k: A Multi-Turn Diffusion Editing Benchmark for Localized AIGC Detection}

\author{%
  Hai Ci$^1$\thanks{Equal Contribution}, \quad Ziheng Peng$^{2\ast}$, \quad Pei Yang$^1$, \quad Yingxin Xuan$^{1}$, \quad Mike Zheng Shou$^1$\thanks{Corresponding Author} \\
  $1$. Show Lab, National University of Singapore\\
  $2$. South China University of Technology\\
  \texttt{\{cihai03,mike.zheng.shou\}@gmail.com} \\
}

\begin{document}
\maketitle
\input{sec/abstract}
\input{sec/introduction}
\input{sec/related_work}
\input{sec/method}

\input{sec/experiment}

\input{sec/conclusion}
{
    \small
    \bibliographystyle{ieeenat_fullname}
    \bibliography{main}
}
\include{sec/X_suppl}


\end{document}

%% file: sec/abstract.tex
\begin{strip}
    \centering
    \includegraphics[width=0.9\linewidth]{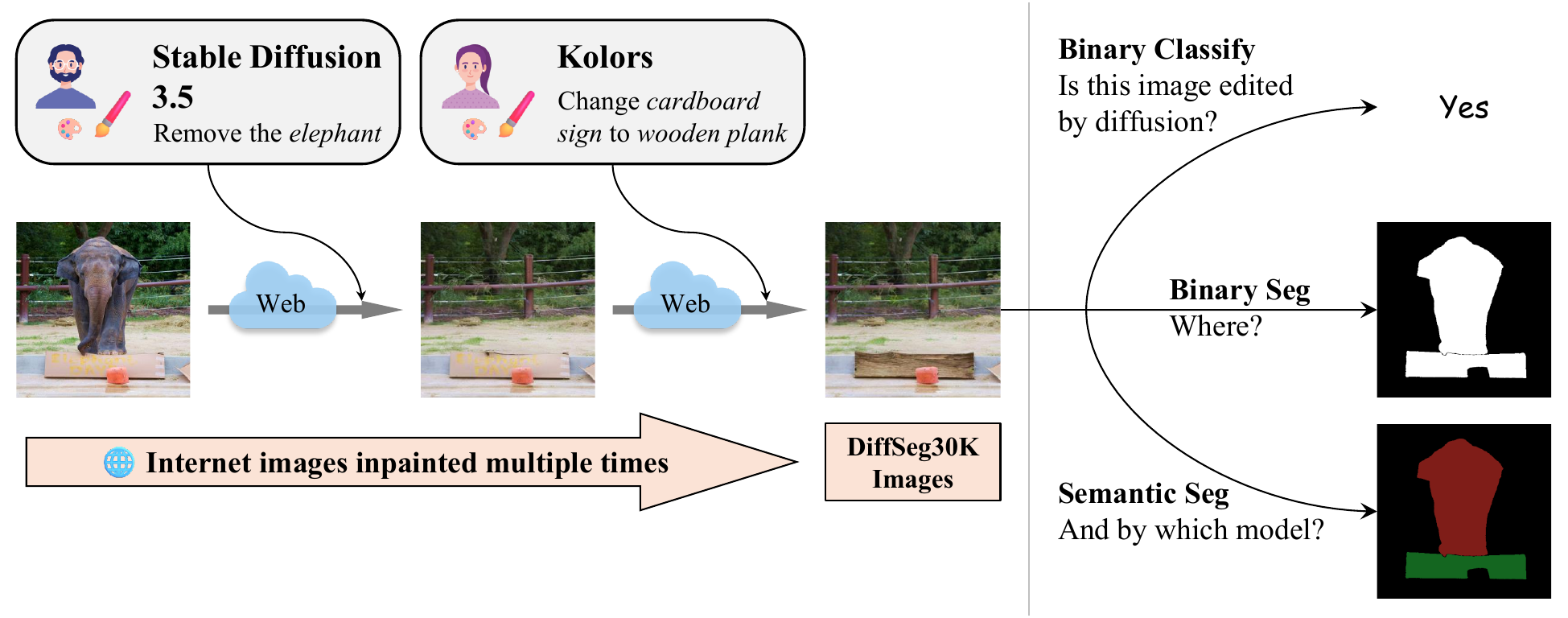}
    \captionof{figure}{\textbf{Motivation.} Images shared online may be independently edited multiple times by different users using different models. DiffSeg30k simulates this real-world scenario through multi-turn diffusion-based edits, enabling novel fine-grained AIGC detection - simultaneous edit localization and model attribution.}
    \label{fig:teaser}
\end{strip}

\begin{abstract}
  Diffusion-based editing enables realistic modification of local image regions, making AI-generated content harder to detect. Existing AIGC detection benchmarks focus on classifying entire images, overlooking the localization of diffusion-based edits.
  We introduce DiffSeg30k, a publicly available dataset of 30k diffusion-edited images with pixel-level annotations, designed to support fine-grained detection. DiffSeg30k features: 1) In-the-wild images—we collect images or image prompts from COCO to reflect real-world content diversity; 2) Diverse diffusion models—local edits using eight SOTA diffusion models; 3) Multi-turn editing—each image undergoes up to three sequential edits to mimic real-world sequential editing; and 4) Realistic editing scenarios—a vision-language model (VLM)-based pipeline automatically identifies meaningful regions and generates context-aware prompts covering additions, removals, and attribute changes.
  DiffSeg30k shifts AIGC detection from binary classification to semantic segmentation, enabling simultaneous localization of edits and identification of the editing models. We benchmark three baseline segmentation approaches, revealing significant challenges in semantic segmentation tasks, particularly concerning robustness to image distortions. Experiments also reveal that segmentation models, despite being trained for pixel-level localization, emerge as highly reliable whole-image classifiers of diffusion edits, outperforming established forgery classifiers while showing great potential in cross-generator generalization.
  We believe DiffSeg30k will advance research in fine-grained localization of AI-generated content by demonstrating the promise and limitations of segmentation-based methods. DiffSeg30k is released at: \url{https://huggingface.co/datasets/Chaos2629/Diffseg30k}
\end{abstract}

%% file: sec/introduction.tex
\section{Introduction}

\input{tables/cmp_different_datasets}

Detecting AI-generated content (AIGC) holds significant societal importance due to its implications for trust, misinformation, privacy, and copyright protection~\citep{wen2023tree,ci2024ringid,ci2024wmadapter,xing2025optmark}. With the advancement of generative AI technologies, particularly diffusion-based editing techniques~\citep{flux,SD3.5,podell2023sdxl,ho2020denoising,bai2025impossible}, it has become possible to produce highly realistic modifications in localized image regions~\citep{meng2021sdedit,Stablediffusion}. Such capabilities raise serious concerns, as these subtle yet realistic edits pose considerable challenges to content verification and digital forensics~\citep{sander2024watermark,simpleavg,ctrlregen,song2025idprotector,song2024anti}.

Current methods and benchmarks for AIGC detection predominantly focus on classifying entire images~\citep{bird2024cifake,zhu2023genimage,defake,verdoliva20222022}, with only limited consideration for localized editing scenarios. This has resulted in a notable gap: the lack of systematic benchmarks specifically designed to evaluate precise detection and localization of AI-generated edits, especially those produced using increasingly popular diffusion-based methods~\citep{SD3.5,flux}. Traditional related tasks, such as Image Forgery Localization (IFL)~\citep{guo2023hierarchical,guillaro2023trufor,wang2022objectformer,zhu2023progressive,liu2022pscc,yu2024diffforensics}, typically address edits made using conventional techniques like copy-move or splicing, yet existing IFL benchmarks rarely cover advanced diffusion-based editing approaches.


In this paper, we introduce \textbf{DiffSeg30k}, a publicly available dataset comprising 30k diffusion-edited images with pixel-level annotations, specifically aimed at facilitating the detection and localization of diffusion-based edits. To reflect realistic, practical scenarios, DiffSeg30k incorporates four key features: 1) \textit{In-the-wild images}—we collect natural images from COCO dataset~\citep{coco} to ensure real-world content diversity. In addition, we leverage COCO-derived prompts with diffusion models to generate complementary AI-based images; 2) \textit{Diverse diffusion models}—local editing is conducted using eight state-of-the-art diffusion models; 3) \textit{Realistic user-driven edits}—an automated annotation pipeline leveraging vision-language models (VLMs)~\citep{Qwen2.5-VL} identifies semantically meaningful regions and generates context-aware editing prompts, including object addition, removal, and attribute modification; and 4) \textit{Multi-turn editing}—each image undergoes up to three sequential edits using different diffusion models, simulating realistic scenarios in which online-shared images are iteratively modified (see~\cref{fig:teaser}). Beyond pixel-level annotations of edited areas, the specific diffusion model used for each edit is annotated, defining a novel semantic segmentation task for fine-grained localization that simultaneously identifies the edited regions and the corresponding editing models. \cref{tab:cmp_different_datasets} compares DiffSeg30k with existing benchmarks containing local editing data.


Leveraging DiffSeg30k, we benchmark three baseline segmentation models—FCN~\citep{long2015fully}, SegFormer~\cite{xie2021segformer}, and Deeplabv3+~\citep{chen2017rethinking,chen2018encoder}—on both binary and semantic segmentation tasks. Our preliminary experiments reveal several key findings: (1) model capacity matters; (2) semantic segmentation in multi-turn editing scenarios remains highly challenging; (3) baseline models are sensitive to post-hoc image transformations such as resizing and JPEG compression; (4) segmentation models can serve as strong whole-image classifiers and largely outperform established classifiers; and (5) segmentation models exhibits strong generalization to unseen generators, indicating promising potential for building more generalizable detection models. We hope the release of DiffSeg30k and these initial results will draw community attention to the important problem of diffusion-based editing localization and attribution, shifting AIGC detection from whole-image classification paradigm to more universe and powerful segmentation paradigm.


%% file: tables/cmp_different_datasets.tex
\begin{table*}[]
\centering
\caption{\textbf{Comparison with previous AIGC detection benchmarks.} We just compare datasets that includes local editing data.}
\label{tab:cmp_different_datasets}
\resizebox{0.85\textwidth}{!}{%
\begin{tabular}{lccccl}
\toprule
\multicolumn{1}{l}{\multirow{2}{*}{\textbf{Datasets}}} & \multirow{2}{*}{\textbf{Image}} & \multirow{2}{*}{\textbf{Availability}} & \multirow{2}{*}{\textbf{\begin{tabular}[c]{@{}c@{}}\# Max Edit\\ Turns\end{tabular}}}  & \multicolumn{2}{c}{\textbf{Diffusion Editing Models}} \\ \cline{5-6} 
\multicolumn{1}{c}{}                                   &                                 &                                                                                      &                                        & \textbf{\# Total} & \multicolumn{1}{c}{\textbf{Name}}        \\ \hline
HIFI-Net~\citep{guo2023hierarchical}                    & General                    & {\color{Green} \ding{51}}      & 1                                                                                                 & 0          & -                                        \\
COCO Glide~\citep{guillaro2023trufor}                   & General                 & {\color{Green} \ding{51}}         & 1                                                                                                 & 1          & Glide~\citep{nichol2021glide}                                    \\
OnlineDet~\citep{epstein2023online}                     & General                 & {\color{red} \ding{55}}        & 1                                                                                                    & 3          & SD1~\citep{Stablediffusion}, SD2~\citep{Stablediffusion}, Adobe Firefly~\citep{firefly}                \\
DA-HFNet~\citep{liu2024hfnet}                           & General                  & {\color{red} \ding{55}}        & 1                                                                                                   & 2          & InpaintAnything~\citep{InpaintAnything}, Paint by Example~\citep{yang2023paint} \\
BR-Gen~\citep{brgen} & General & {\color{Green} \ding{51}} & 1 & 3 & SDXL~\citep{podell2023sdxl}, BrushNet~\citep{ju2024brushnet}, PowerPaint~\citep{powerpaint} \\
WFake~\citep{wfake}                                     & Facial                      & {\color{Green} \ding{51}}        & 1                                                                                              & 2          & Repaint-p2~\citep{lugmayr2022repaint}, Repaint-LDM~\citep{Stablediffusion}                  \\
\hline 
\cellcolor{lightblue!70} DiffSeg30k(Ours)                                                   & \cellcolor{lightblue!70} General                  & \cellcolor{lightblue!70} {\color{Green} \ding{51}}         & \cellcolor{lightblue!70} 3                                                                                                & \cellcolor{lightblue!70} 8          & \cellcolor{lightblue!70} \begin{tabular}[l]{@{}l@{}}SD2~\citep{Stablediffusion}, SD3.5~\citep{SD3.5}, SDXL~\citep{podell2023sdxl}, \\ Flux.1~\citep{flux}, Glide~\citep{nichol2021glide}, Kolors~\citep{kolors}, \\ HunyuanDiT1.1~\citep{li2024hunyuandit}, Kandinsky 2.2~\citep{kandinsky2}\end{tabular}\\ \bottomrule
\end{tabular}%
}
\end{table*}

%% file: sec/related_work.tex
\section{Related Work}
\subsection{Image Forgery Localization}
This task focuses on detecting image regions modified by traditional editing techniques such as copy-move and splicing. Representative benchmarks include CASIA~\citep{dong2013casia}, Columbia~\citep{ng2009columbia}, NIST16~\citep{nist16}, IMD20~\citep{novozamsky2020imd2020}, Coverage~\citep{wen2016coverage}, FantasticReality~\citep{fantasticreality}, PSCC-Net~\citep{liu2022pscc}, OpenForensics~\citep{le2021openforensics}, etc. Based on these datasets, various approaches have been developed, leveraging transformer-based feature extraction~\citep{guillaro2023trufor,wang2022objectformer}, progressive learning~\citep{zhu2023progressive,liu2022pscc}, hierarchical mechanisms~\citep{guo2023hierarchical}, and diffusion priors~\citep{yu2024diffforensics} to localize edits.

\begin{figure*}[t]
    \centering
    \includegraphics[width=0.8\linewidth]{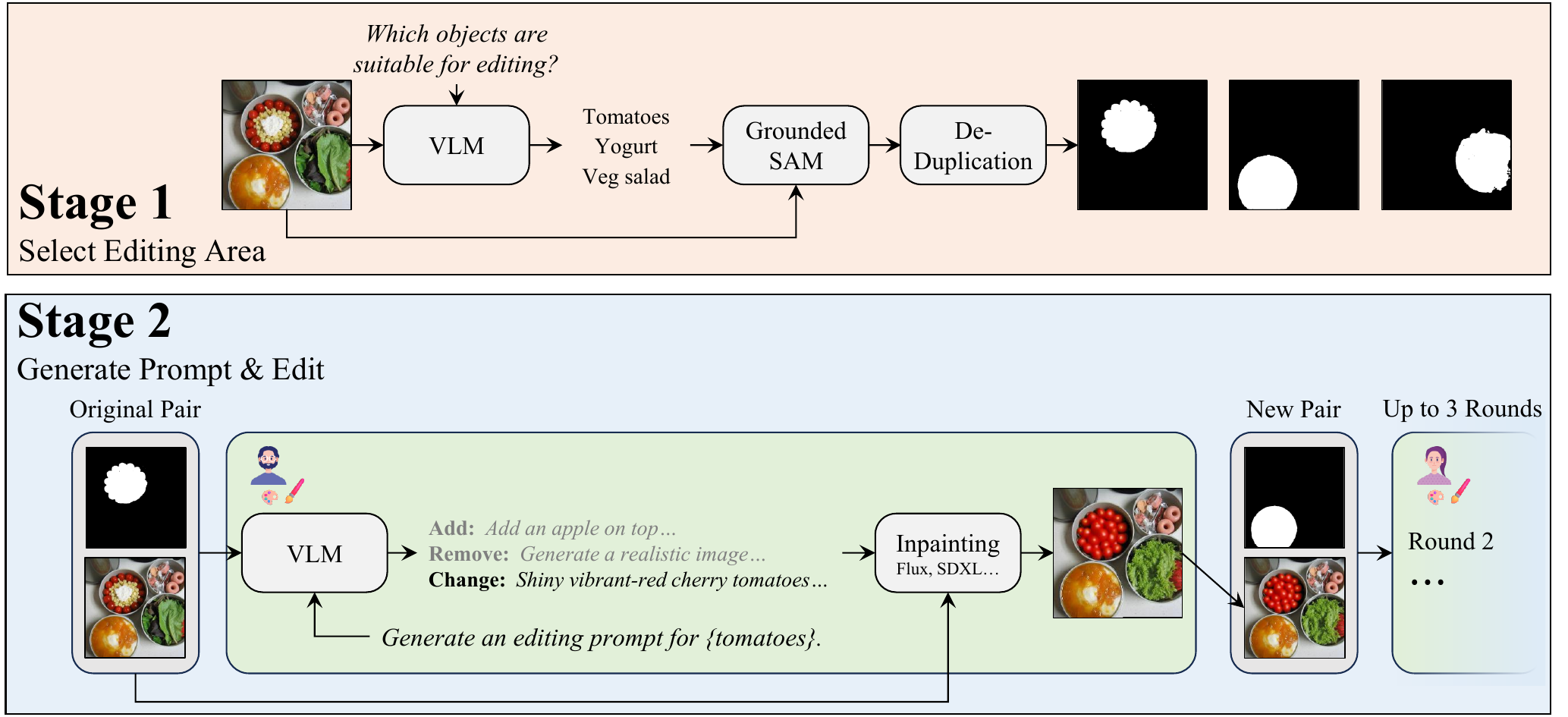}
    \caption{\textbf{Automatic data collection framework.} The pipeline consists of two stages: (1) identifying editable regions using VLMs~\citep{Qwen2.5-VL} and Grounded-SAM~\citep{groundedsam} to generate object masks; (2) generating context-appropriate editing prompts with VLMs and applying sequential diffusion-based edits to the selected regions.}
    \label{fig:framework}
\end{figure*}

\subsection{AIGC Detection}
The task of AIGC detection involves identifying content created by generative AI models such as GANs and diffusion models. Current datasets~\citep{zhu2023genimage,defake,bird2024cifake,verdoliva20222022,deepart,he2021forgerynet,guo2023hierarchical,miao2025ddl} and methods~\citep{cozzolino2024raising,cnnspot,zhong2023patchcraft,cozzolino2024zero,wang2025scaling,jiang2025edittrack,sun2025fragfake} predominantly focus on classifying entire images. Existing datasets can be categorized into two types according to their image content: facial images~\citep{he2021forgerynet,yang2019exposing,wang2019fakespotter,dang2020detection,gandhi2020adversarial,miao2025ddl} and general images~\citep{zhu2023genimage,defake,bird2024cifake,verdoliva20222022,deepart}. 
Several datasets~\citep{guillaro2023trufor,wfake,epstein2023online,liu2024hfnet} include limited numbers ($\leq 3$) of diffusion models and only involving single local edit per image,  as detailed in~\cref{tab:cmp_different_datasets}. In contrast, our proposed DiffSeg30k dataset focuses on general images, covers eight different diffusion models, and incorporates multi-turn editing by different models on each image. This comprehensive dataset provides systematic support for more general semantic segmentation tasks beyond conventional first-classify-then-binary-segmentation paradigm~\cite{guo2023hierarchical,yan2025gamma} used in single-turn editing datasets.

%% file: sec/method.tex
\section{DiffSeg30k benchmark}
\label{sec:dataset}

\begin{figure*}[t]
    \centering
    \includegraphics[width=0.9\linewidth]{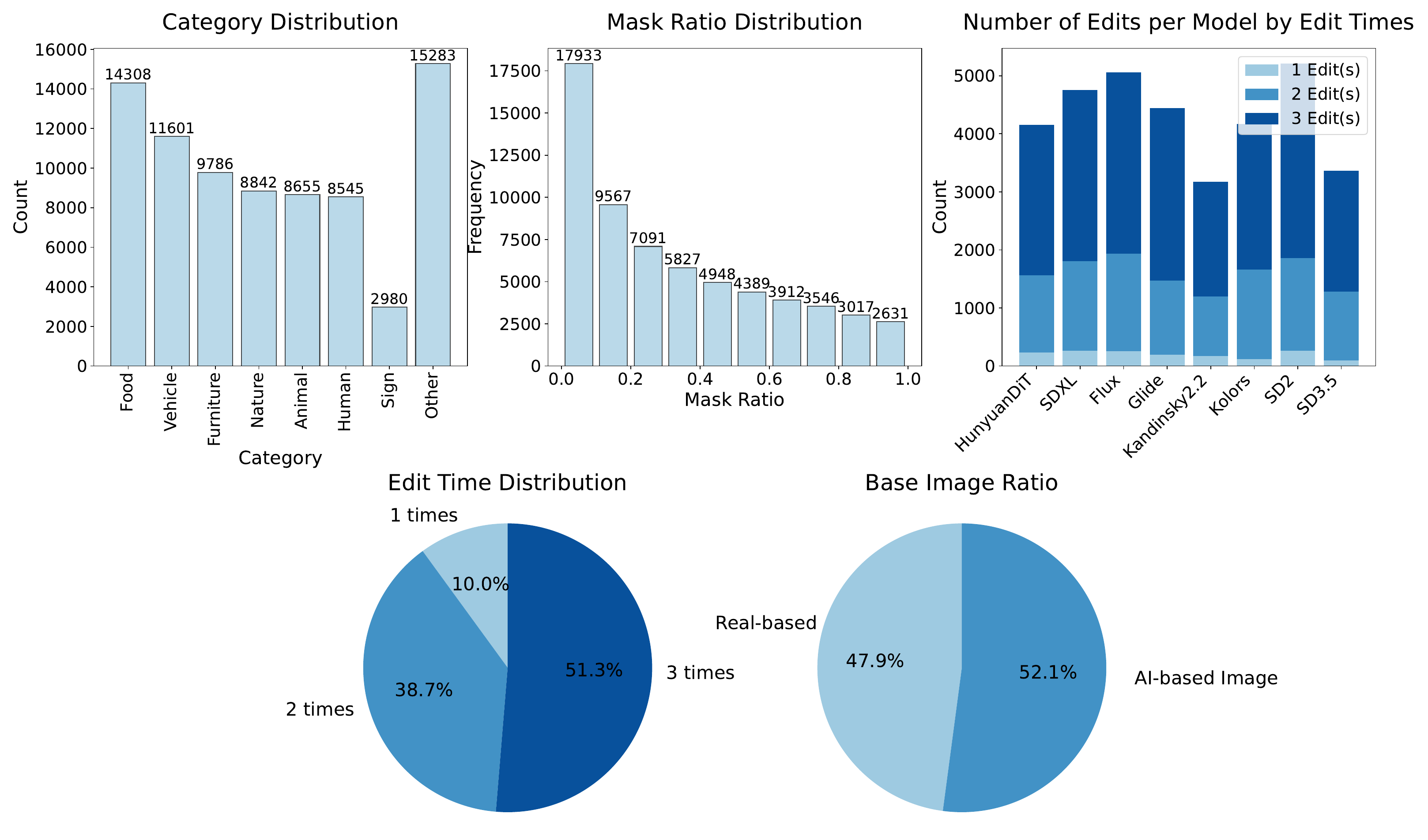}
    \caption{\textbf{DiffSeg30k Statistics}. The distributions reflect our balanced data collection strategy: (a) Object category distribution shows Food (14,308) are most frequently edited, with an overall balanced distribution across categories. (b) Mask area ratio distribution reveals a decreasing trend from smaller to larger regions. (c) Editing count per model indicates uniform participation of all eight diffusion models across first, second, and third editing rounds. (d) We manually increased the frequency of multi-turn editing, with the first, second, and third rounds occurring at an approximate 1:4:5 ratio, as multi-turn editing subsumes single-turn cases and presents a more challenging scenario. (e) The base image ratio is approximately balanced between real and AI-generated images.}
    \label{fig:stats}
\end{figure*}

\subsection{Dataset construction}
We developed an automated image editing and annotation pipeline based on vision-language models (VLM), illustrated in~\cref{fig:framework}. The pipeline comprises two stages: 1) generating candidate editing regions and 2) creating suitable editing prompts for these regions, followed by image editing.

\noindent\textbf{Stage 1—generating candidate editing regions.} As shown in~\cref{fig:framework}, given an arbitrary image, we first use Qwen2.5-VL~\citep{Qwen2.5-VL}  to identify distinct objects present in the image. The identified object categories are then provided to Grounded-SAM~\citep{groundedsam} to generate corresponding masks for each object. Subsequently, we calculate the Intersection-over-Union (IoU) among object masks and eliminate redundant masks with IoU greater than 70\% to avoid repetitive editing of the same object. Finally, we randomly select between one to three object masks for editing.

\noindent\textbf{Stage 2—editing prompt generation \& inpainting.} As depicted in~\cref{fig:framework}, each selected object mask from Stage 1 is paired with the original image and fed into the Qwen2.5-VL model, prompting it to generate contextually appropriate editing prompts. For inpainting, the generated editing prompt, corresponding object mask, and the original image are simultaneously provided to a diffusion model to perform localized inpainting. For multi-turn editing, the edited image can form a new pair with the remaining object masks, repeating the prompt generation and inpainting processes. Thus, we obtain images edited sequentially up to three times, depending on the number of masks selected in Stage 1.

\noindent\textbf{Data diversity \& balance.} To ensure the diversity and balance of the edited images, we specifically considered the following six aspects:
1) \textit{Balanced Candidate Object Types:} We prompted the VLM to prioritize selecting humans, as human-centric edits are significant in practical applications, yet VLMs tend to under-select humans naturally.
2) \textit{Balanced Candidate Mask Area:} We encouraged VLM to prioritize larger objects for editing, counteracting the empirical tendency of VLM to select smaller objects.
3) \textit{Balanced Edit Types:} During editing prompt generation, we prompted VLM to randomly choose from three edit types: attribute changes, object additions, and object removals. For object additions, we followed the approach of SEED-Data-Edit~\citep{ge2024seed}, first removing an object and then adding a new object in the same position to ensure natural placement.
4) \textit{Balanced Edit Models:} For inpainting, we randomly selected one diffusion model from eight state-of-the-art diffusion models to execute an editing prompt.
5) \textit{Enhanced Multi-turn Editing:} We manually increased the frequency of multi-turn editing, with the first, second, and third turns  occurring at an approximate 1:4:5 ratio, as multi-turn editing subsumes single-turn cases and presents a more challenging scenario.
6) \textit{Balanced Real and AI Bases:} About half of the base images were real, collected from COCO, while the other half were AI-generated using COCO prompts with random diffusion models.

\subsection{Sanity check}
Our automated pipeline occasionally produces low-quality edits due to (i) limited diffusion editing capability (e.g., incomplete object removal) or (ii) mask errors from Grounded-SAM~\citep{groundedsam}. 
These issues largely stem from our framework’s ambition to simulate diverse and realistic editing scenarios. In contrast, existing datasets such as OnlineDet~\citep{epstein2023online} rely on random masks with potentially empty prompts, while wFake~\citep{wfake} focuses only on predefined facial regions (e.g., eyes, nose). Our approach covers a broader and more complex range of edits, though at the cost of occasional artifacts.
To improve dataset quality, we employ Qwen2.5-VL~\citep{Qwen2.5-VL} to assign image-level quality scores (0–5) and automatically discard edits with severe artifacts (score $<$ 3). Details of filtering are provided in supplementary.
    
\subsection{Dataset statistics}
We conducted a statistical analysis of DiffSeg30k, illustrated in~\cref{fig:stats}. 
It reveals that smaller objects are edited more frequently, whereas larger objects tend to be edited less often, with editing frequency smoothly decreasing as object area increases. Objects such as food, vehicles, furniture, nature scenery, animals, humans, and signs are among the most frequently edited. The dataset maintains a balanced distribution across editing models and base image distribution. \cref{fig:examples} shows some editing results.

%% file: sec/experiment.tex
\section{Benchmarking baselines}
\label{sec:exp}

\input{tables/exp_main}

\input{tables/exp_robustness}

\begin{figure*}[t]
    \centering
    \includegraphics[width=0.9\linewidth]{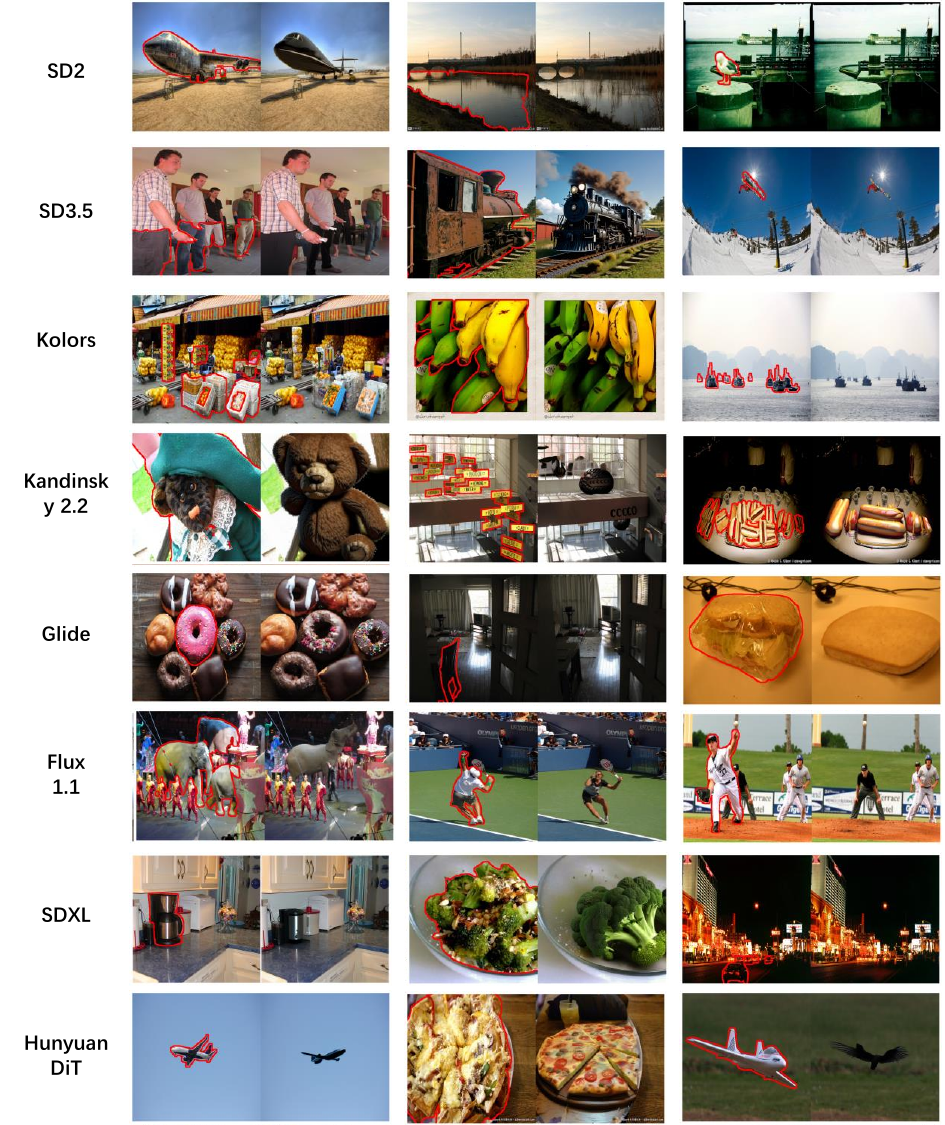}
    \caption{\textbf{Example edited images.} For each image pair, the left shows the original image with red contours marking the regions to be edited, and the right presents the corresponding editing result.}
    \label{fig:examples}
\end{figure*}

\begin{figure*}[t]
    \centering
    \includegraphics[width=1\linewidth]{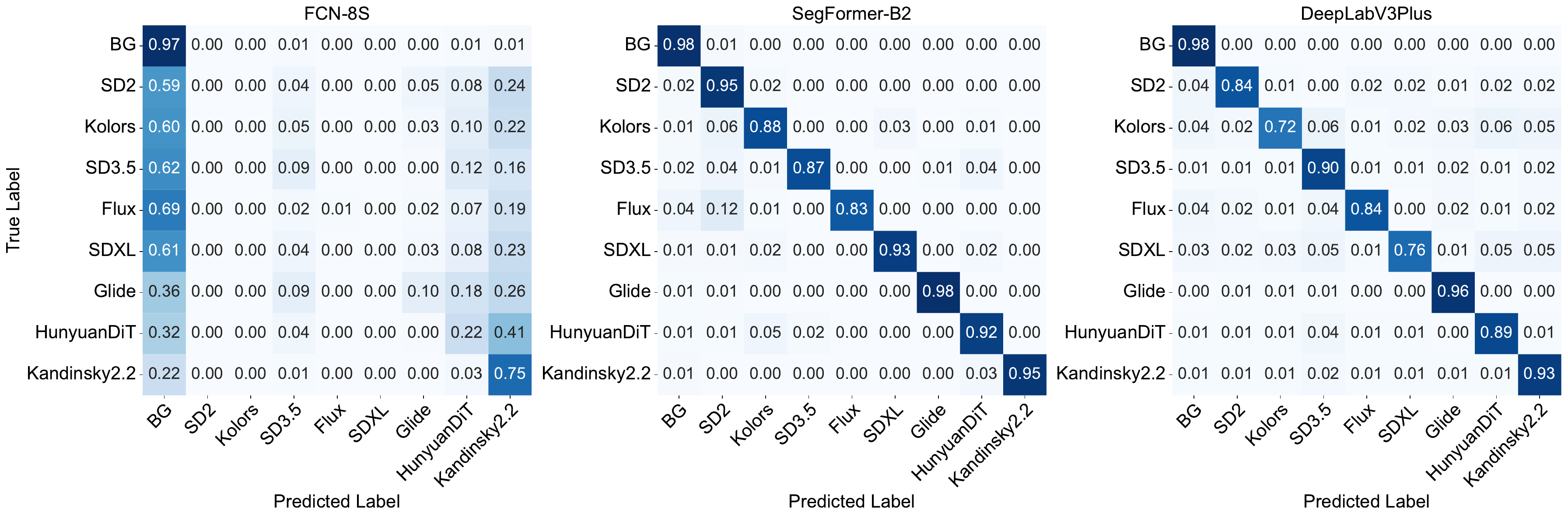}
    \caption{\textbf{Confusion matrix on the semantic segmentation task.}}
    \label{fig:exp_confusion}
\end{figure*}

\subsection{Experimental setup}
\textbf{Editing data preparation.} To match the input requirements of diffusion editing models as well as achieving high editing quality, COCO images are first resized so that the shorter side is 1024 pixels, followed by a center crop to 1024×1024. For model-specific compatibility, images are resized to 256 for Glide during editing. 

\noindent\textbf{Training details.} Due to the lack of existing baseline segmentation models on general AIGC localization to handle semantic segmentation, we trained two CNN baselines based on classic semantic segmentation frameworks: FCN-8s~\citep{long2015fully} and Deeplabv3+\citep{chen2017rethinking,chen2018encoder} with ResNet50\citep{he2016deep} and 1 transfomer baseline SegFormer-B2~\cite{xie2021segformer}.  We evaluated these models on two tasks: binary segmentation (localizing edited areas) and semantic segmentation (localizing edited areas while distinguishing among different editing models). The dataset is split into training and validation sets in an 8:2 ratio, and all results are reported on the validation set. 
Models are trained on 512×512 resolution (for acceleration) using default hyperparameters from open-source implementations. We additionally include 5k unedited real COCO images in the training set to help suppress false positives on real images. Training FCN-8s, SegFormer-B2, and Deeplabv3+ takes approximately 8, 6, and 3 hours, respectively, on an NVIDIA 4090 GPU.

\noindent\textbf{Evaluation metric.} To evaluate classification performance, we report accuracy (Acc) and mean average precision (mAP). For segmentation, we report pixel-wise accuracy (Acc), mask Intersection-over-Union (mIoU) and boundary-F1 score (bF1) between predicted and ground truth masks. The background class is excluded from mIoU and bF1 calculations to specifically reflect models' abilities to detect edited areas.

\subsection{Results}
\noindent\textbf{Model capacity matters.}
As shown in~\cref{tab:exp_main}, the older FCN-8s struggles with both binary and semantic localization of diffusion-edited regions. In contrast, more powerful segmentation models Deeplabv3+ and SegFormer , performs substantially better, achieving strong binary localization with an mIoU of 0.974 and 0.961. These results highlight the importance of model representational capacity. 

\begin{figure*}[ht]
    \centering
    \includegraphics[width=0.9\linewidth]{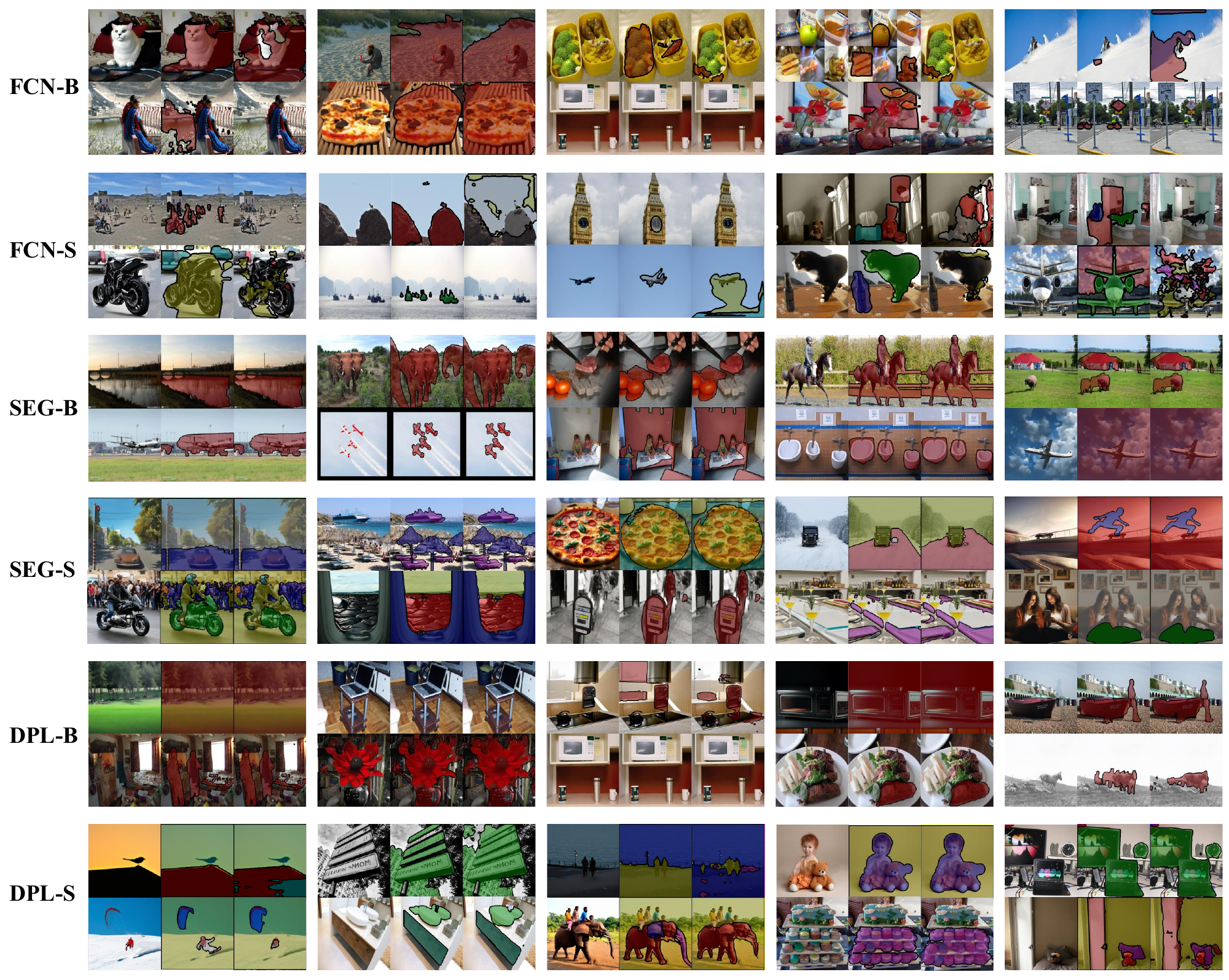}
    \caption{\textbf{Segmentation results.}  ``SEG'' and ``DPL'' denote SegFormer and Deeplab v3+, respectively.  ``-B'' and ``-S'' refer to binary and semantic segmentation, respectively. In each image group, the first column shows edited images, the second column shows the ground truth masks, and the third column shows the predicted masks.
    In each mask,
\textcolor{BrickRed}{Crimson} corresponds to \textcolor{BrickRed}{Stable Diffusion 2},
\textcolor{ForestGreen}{ForestGreen} corresponds to \textcolor{ForestGreen}{Kolors},
\textcolor{GreenYellow}{Yellow} corresponds to \textcolor{GreenYellow}{Stable Diffusion 3.5 Medium},
\textcolor{RoyalBlue}{RoyalBlue} corresponds to \textcolor{RoyalBlue}{Flux},
\textcolor{Purple}{MediumPurple} corresponds to \textcolor{Purple}{Stable Diffusion XL 1.0},
\textcolor{TealBlue}{TealBlue} corresponds to \textcolor{TealBlue}{Glide},
\textcolor{Gray}{Gray} corresponds to \textcolor{Gray}{Hunyuan-DiT},
and \textcolor{red}{red} corresponds to \textcolor{red}{Kandinsky 2.2}.}
    \label{fig:exp_seg_examples}
\end{figure*}

\noindent\textbf{Semantic segmentation is challenging.}
Simultaneously localizing edited regions and attributing them to the correct diffusion model is substantially more difficult than binary localization. Even for the best performed model SegFormer, the mIoU for semantic segmentation drops to 0.825, leaving significant room for improvement. For Deeplabv3+ and FCN, this number further drops significantly to 0.76 and 0.23. \cref{fig:exp_confusion} shows the confusion matrices: FCN-8s tends to misclassify edits from all models as background pixels, while SegFormer struggles particularly with  Flux. Deeplab v3+ struggles with Kolors and SDXL. It also often misclassifies SD2 and Flux. \cref{fig:exp_seg_examples} further illustrates representative results. FCN-8s produces largely random outputs in both tasks, whereas SegFormer and Deeplab v3+ achieves more coherent segmentations but occasionally misclassifies the editing model. In the following detailed ablation, we only evaluate Deeplabv3+ and SegFormer.



\vspace{4pt}
\noindent\textbf{Baseline models are sensitive to image transformations.}
We evaluated the robustness of SegFormer and DeepLabv3+ under common image transformations, including JPEG compression and resizing. As shown in~\cref{tab:exp_robustness}, image transformations substantially degraded the performance of both Deeplabv3+ and SegFormer, with severe mIoU drops—semantic segmentation almost failed under JPEG compression. This highlights the need for improved network architectures or data augmentation strategies to achieve deployable localization models. Interestingly, DeepLabv3+ exhibits partial resilience in binary segmentation settings, while SegFormer demonstrates certain robustness when resized to higher resolutions.



\noindent\textbf{Segmentation models achieve strong cross-generator generalization.}
\cref{tab:exp_generalization} presents the cross-generator generalization results of SegFormer and Deeplabv3+. Training was conducted on images edited by six diffusion models and tested on edits from two unseen models, with three different combinations evaluated. Both SegFormer and Deeplabv3+ demonstrate strong cross-generator generalization, with mIoU consistently above 0.9 and 0.8, respectively. Notably, both models were not specifically designed for such generalization, suggesting that segmentation-based methods hold great potential for transferable AIGC detection.

\input{tables/exp_generalization}
\input{tables/exp_classification}
\input{tables/exp_cls_gen}

\vspace{4pt}
\noindent\textbf{Segmentation models as strong generalizable whole-image classifiers.}
Given the strong performance of SegFormer and DeepLabv3+ in binary segmentation, we further investigate whether they can serve as powerful whole-image AIGC classifiers. Unlike standard AIGC classification, which determines whether an entire image is generated by a diffusion model, our task is more challenging: models must identify whether diffusion-edited regions exist within an image. 
We set a threshold such that if the non-background mask area exceeds a minimal threshold 1\% of the image area, the image is considered to contain edited content. We compare with two established AIGC classification models, CNNSpot~\citep{cnnspot} and UniversalFakeDet~\citep{unifakedet}. During training, we additionally collect 30k real images from COCO~\citep{coco} as negative samples to train the classification baselines.
Results in~\cref{tab:exp_cls} reveal a striking observation: when repurposed for whole-image classification, segmentation models outperform dedicated AIGC classifiers by a significant margin. We attribute this advantage to the fine-grained, per-pixel supervision employed during segmentation training. Our dataset includes images containing multiple, interleaved edited regions from different generators, allowing segmentation models to develop a deeper understanding of localized generative cues. Notably, although UniversalFakeDet was originally designed for cross-model generalization in whole-image classification, it performs markedly worse than general-purpose segmentation models in cross-model evaluation (\cref{tab:exp_cls_gen}).  Cross-model generalization has long been a challenge in the AIGC classification community. Through this experiment, we demonstrate that segmentation models trained on multi-turn edited data exhibit strong potential to address this issue, suggesting a paradigm shift from AIGC classification toward more robust AIGC segmentation.

%% file: tables/exp_main.tex

\begin{table*}
\centering
\caption{\textbf{Performance of baseline segmentation models.} Results of SegFormer and Deeplabv3+ on binary segmentation (localizing edited regions) and semantic segmentation (localizing and attributing edits to specific diffusion models).}
\label{tab:exp_main}
\begin{tabular}{lllllll}
\toprule
& \multicolumn{3}{l}{Binary segmentation} & \multicolumn{3}{l}{Semantic segmentation} \\ \cline{2-7}  & Acc           & mIoU           & bF1 &  Acc             & mIoU   & bF1         \\ 
\hline
SegFormer~\citep{xie2021segformer}      & \textbf{0.989}          & 0.961   & \textbf{0.678}      & \textbf{0.952}           & \textbf{0.825}    & \textbf{0.532}      \\
Deeplab v3+~\citep{chen2017rethinking} & 0.984          & \textbf{0.974}   & 0.761      & 0.916           & 0.760     & 0.431     \\
FCN-8s~\citep{long2015fully} & 0.838          & 0.699   & 0.457      & 0.331           & 0.203     & 0.024     \\
\bottomrule
\end{tabular}
\end{table*}

%% file: tables/exp_robustness.tex
\begin{table*}
\centering
\caption{\textbf{Robustness to image transformations.} We evaluate DeepLabv3+and SegFormer under JPEG compression with varying transformations to assess their sensitivity to common post-processing operations. \textcolor{gray}{Baseline} indicates no distortions.}
\label{tab:exp_robustness}
\begin{tabular}{llllllll}
\toprule
                             &             & \multicolumn{3}{l}{Binary segmentation}               & \multicolumn{3}{l}{Semantic segmentation}             \\ \cline{3-8} 
                             & Distortion  & \multicolumn{1}{l}{Acc}   & \multicolumn{1}{l}{mIoU} & \multicolumn{1}{l}{bF1}  & \multicolumn{1}{l}{Acc}   & \multicolumn{1}{l}{mIoU} & \multicolumn{1}{l}{bF1}  \\ \hline
\multirow{5}{*}{SegFormer~\citep{xie2021segformer}}      &  \textcolor{gray}{Baseline}          &  \multicolumn{1}{l}{\textcolor{gray}{0.989}} &  \multicolumn{1}{l}{\textcolor{gray}{0.961}} & \multicolumn{1}{l}{\textcolor{gray}{0.678}} &  \multicolumn{1}{l}{\textcolor{gray}{0.953}} &  \multicolumn{1}{l}{\textcolor{gray}{0.825}} & \multicolumn{1}{l}{\textcolor{gray}{0.532}}\\
                             & JPEG 60      & 0.705                     & 0.260      & 0.257                 & 0.665                     & 0.032     & 0.197                 \\
                             & JPEG 80      & 0.775                     & 0.291         & 0.338              & 0.724                     & 0.188         & 0.305              \\
                             & resize 256  & 0.565                     & 0.360     & 0.286                 & 0.616                     & 0.299     & \textbf{0.338}                 \\
                             & resize 1024 & \textbf{0.966}                     & \textbf{0.898}          & \textbf{0.548}             & \textbf{0.821}                     & \textbf{0.414}          & 0.296             \\ 
                             \bottomrule 
\multirow{5}{*}{Deeplab v3+~\citep{chen2017rethinking}}      &  \textcolor{gray}{Baseline}          &  \multicolumn{1}{l}{\textcolor{gray}{0.984}} &  \multicolumn{1}{l}{\textcolor{gray}{0.974}} & \multicolumn{1}{l}{\textcolor{gray}{0.761}} &  \multicolumn{1}{l}{\textcolor{gray}{0.916}} &  \multicolumn{1}{l}{\textcolor{gray}{0.760}} & \multicolumn{1}{l}{\textcolor{gray}{0.431}}\\
                             & JPEG 60      & 0.867                     & 0.782      & 0.556                 & 0.184                     & 0.013     & 0.019                 \\
                             & JPEG 80      & \textbf{0.907}                     & \textbf{0.847}         & 0.613              & \textbf{0.190}                     & 0.014         & 0.019              \\
                             & resize 256  & 0.811                     & 0.751     & \textbf{0.631}                 & 0.021                     & 0.012     & \textbf{0.021}                  \\
                             & resize 1024 & 0.816                     & 0.729          & 0.499             & 0.168                     & \textbf{0.168}          & 0.014             \\ 
                             \bottomrule
       
\end{tabular}
\end{table*}

%% file: tables/exp_generalization.tex
\begin{table}[]
\centering
\caption{\textbf{Generalization to unseen diffusion generators}. Segmentation models are trained on images edited by 6 diffusion generators and evaluated on edits from 2 unseen generators to assess cross-generator generalization. \textcolor{gray}{Baseline} indicates training and testing on 8 generators.}
\label{tab:exp_generalization}
\resizebox{\linewidth}{!}{
\begin{tabular}{lllll}
\toprule
                             &                 & \multicolumn{3}{c}{Binary segmentation} \\ \cline{3-5} 
                             & Generalize to & Acc                & mIoU & bF1                \\ \hline
\multirow{4}{*}{SegFormer~\citep{xie2021segformer}}      & \textcolor{gray}{Baseline}               & \textcolor{gray}{0.989}             & \textcolor{gray}{0.961}  &  \textcolor{gray}{0.678}          \\

                             & Flux\_Hunyuan   & \textbf{0.935}             & \textbf{0.930}    & 0.491          \\
                             & SD2\_SD3.5      & 0.927             & \textbf{0.926}    & \textbf{0.497}          \\
                             & Glide\_Kolors   & \textbf{0.957}             & \textbf{0.951}    & 0.578          \\ 
\bottomrule

\multirow{4}{*}{Deeplab v3+~\citep{chen2017rethinking}}      & \textcolor{gray}{Baseline}               & \textcolor{gray}{0.984}             & \textcolor{gray}{0.974}  &  \textcolor{gray}{0.761}          \\

                             & Flux\_Hunyuan   & 0.920             & 0.867    & \textbf{0.526}          \\
                             & SD2\_SD3.5      & \textbf{0.931}             & 0.833    & 0.372          \\
                             & Glide\_Kolors   & 0.953             & 0.927    & \textbf{0.648}          \\ 
\bottomrule
\end{tabular}
}
\end{table}

%% file: tables/exp_classification.tex
\begin{table}[]
\centering
\caption{\textbf{Binary classification results of estabilished AIGC detectors.} SegFormer and Deeplabv3+ are adapted to classification by thresholding.}
\label{tab:exp_cls}
\begin{tabular}{lll}
\toprule
                 & Acc    & mAP    \\ \hline
CNNSpot~\citep{cnnspot}          & 0.942  & 0.979  \\
UniversalFakeDet~\citep{unifakedet} & 0.860 & 0.934 \\ 
\hline
SegFromer ~\citep{xie2021segformer} & \textbf{0.997} & \textbf{0.999} \\ 
Deeplab v3+~\citep{chen2017rethinking} & 0.980 & 0.994 \\ 
\bottomrule
\end{tabular}
\end{table}

%% file: tables/exp_cls_gen.tex
\begin{table}[]
\centering
\caption{\textbf{Cross-generator generalization of differnet classifiers.} The model is trained on images edited by 6 diffusion generators and evaluated on images edited by 2 unseen diffusion generators to assess its ability to generalize across different editing sources. \textcolor{gray}{Baseline} indicates training and testing on 8 generators. Segmentation models are adapted by thresholding.}
\label{tab:exp_cls_gen}
\begin{tabular}{llll}
\toprule
                                  & Generalize to & Acc                       & mAP                       \\ \hline
\multirow{4}{*}{UniFakeDet~\citep{unifakedet}} & \textcolor{gray}{Baseline}      & \multicolumn{1}{r}{\textcolor{gray}{0.860}} & \multicolumn{1}{r}{\textcolor{gray}{0.934}} \\
                                  & Flux\_Hunyuan & 0.768                     & 0.877                     \\
                                  & SD2\_SD3.5    & 0.714                     & 0.817                     \\
                                  & Glide\_Kolors & 0.749                     & 0.844                     \\ \bottomrule
\multirow{4}{*}{SegFromer~\citep{xie2021segformer}} & \textcolor{gray}{Baseline}      & \multicolumn{1}{r}{\textcolor{gray}{\textbf{0.997}}} & \multicolumn{1}{r}{\textcolor{gray}{\textbf{0.999}}} \\
                                  & Flux\_Hunyuan & \textbf{0.978}                     & \textbf{0.986}                     \\
                                  & SD2\_SD3.5    & \textbf{0.982}                     & \textbf{0.986}                     \\
                                  & Glide\_Kolors & \textbf{0.987}                     & \textbf{0.996}                     \\ \bottomrule   
\multirow{4}{*}{Deeplab V3+~\citep{chen2017rethinking}} & \textcolor{gray}{Baseline}      & \multicolumn{1}{r}{\textcolor{gray}{0.980}} & \multicolumn{1}{r}{\textcolor{gray}{0.994}} \\
                                  & Flux\_Hunyuan & 0.935                     & 0.967                     \\
                                  & SD2\_SD3.5    & 0.942                     & 0.965                     \\
                                  & Glide\_Kolors & 0.964                     & 0.985                     \\ \bottomrule              
                                 
\end{tabular}
\end{table}

%% file: sec/conclusion.tex
\section{Conclusion}
\label{sec:conclusion}
We introduce DiffSeg30k, a multi-turn diffusion editing dataset that provides a robust foundation for systematically studying diffusion-based editing localization and model attribution. Through extensive experimental analyses, we highlight both the strengths and current limitations of segmentation methods, laying the groundwork for further research in robust AIGC localization techniques.
\noindent\textbf{License.} This dataset is derived from COCO and is distributed under the COCO license without additional restrictions.

%% file: sec/X_suppl.tex
\clearpage
\maketitlesupplementary
\section{Relation to global editing}
A line of work focuses on \emph{global image editing}~\cite{tan2025ominicontrol,tan2025ominicontrol2,nanobanana2025,wei2024omniedit,cocoinpaint}, where the entire output image is synthesized from scratch. From the perspective of AIGC detection, these methods can be grouped together with standard (“vanilla”) image generation approaches, as they lack the notion of a \emph{local editing region}—all pixels are fully regenerated.  

In contrast, our dataset is intentionally designed around \emph{localized inpainting–style editing}, motivating a shift in AIGC detection from whole-image classification to per-pixel attribution, a more general and fine-grained paradigm. This formulation also lays a foundation for improving global-editing detection, as our experiments show that segmentation models can serve as strong whole-image classifiers.

\section{Do LoRAs affect localization?}
Diffusion models can be enhanced with Low-Rank Adaptation (LoRA) modules~\citep{hu2022lora} to improve efficiency~\citep{ren2024hypersd} or expressiveness~\citep{zhang2025easycontrol,song2024processpainter}. In this experiment, we investigate whether the use of LoRAs impacts the localization performance of baseline segmentation models. Specifically, we apply the Hyper-SD LoRA~\citep{ren2024hypersd} (8 steps, DDIM scheduler~\citep{song2020denoising}) to SDXL~\citep{podell2023sdxl}, which reduces the number of inference steps from 50 to 8. We re-edit all regions in the validation set that were previously modified by SDXL using the LoRA-accelerated version (SDXL-LoRA). Then, we re-evaluate the baseline semantic segmentation model (trained on 8 vanilla generators from the main text, which just saw SDXL-edited images during training). In this setup, edits made by SDXL-LoRA are considered correctly predicted if classified as SDXL, allowing us to evaluate the segmentation model’s generalization ability to LoRA-based variants. As shown in~\cref{tab:supp_lora}, the segmentation model exhibits strong generalization to LoRA variants, with only a minor mIoU drop of approximately 0.03.

\input{tables/supp_lora}

\section{Quality assessment}
\label{sec:supp_quality_filter}
\begin{figure*}[t]
    \centering
    \includegraphics[width=\linewidth]{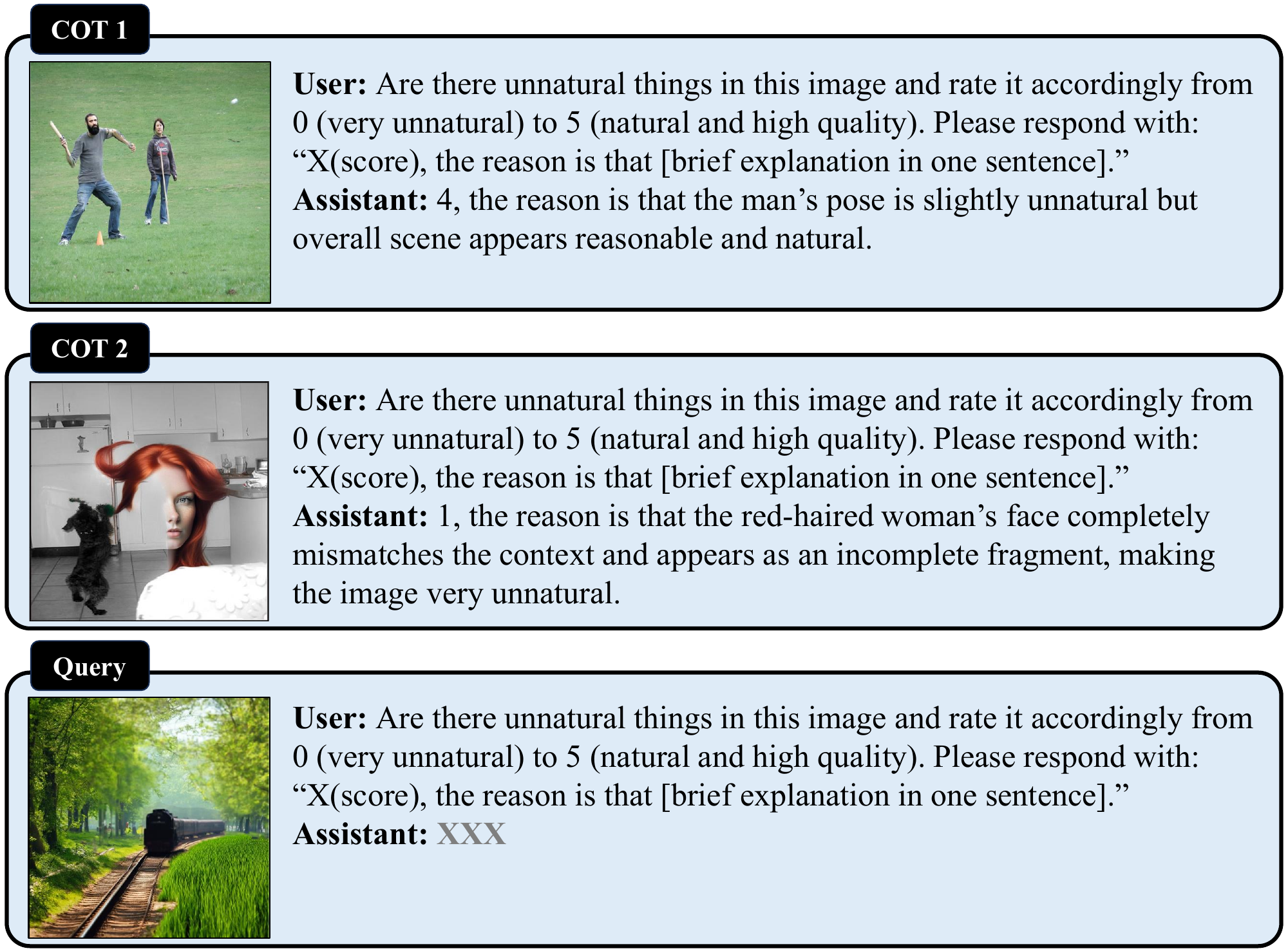}
    \caption{\textbf{Chain-of-thought examples for image quality assessment.}}
    \label{fig:supp_quality_filter}
\end{figure*}

To remove severely low-quality outputs generated by the automated editing framework, we employ Qwen2.5-VL~\citep{Qwen2.5-VL} to evaluate the quality of each edited image. \cref{fig:supp_quality_filter} shows two chain-of-thought demonstration examples provided to Qwen2.5-VL. Images receiving a score below 3 are discarded, resulting in the removal of roughly 50\% of the generated samples. This threshold is intentionally strict, leading to a high rejection rate, but it ensures a higher-quality resulting dataset.

\begin{figure*}[t]
    \centering
    \includegraphics[width=\linewidth]{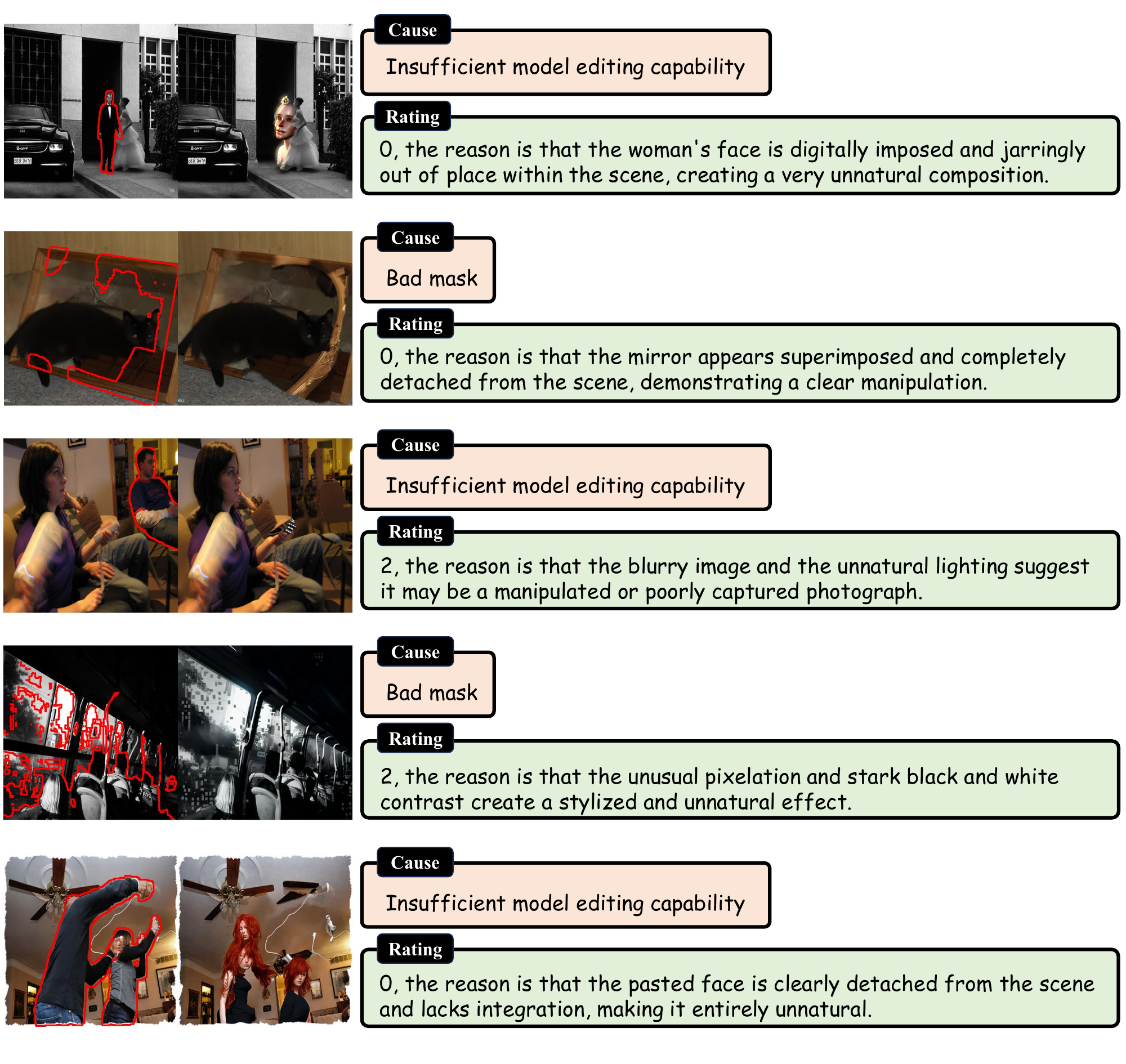}
    \caption{\textbf{Typical low quality edits and VLM ratings.}}
    \label{fig:supp_bad_case}
\end{figure*}

\cref{fig:supp_bad_case} shows typical low-quality edits, usually caused by insufficient diffusion editing capability or inaccurate masks. In the third example, we observe that Qwen2.5-VL occasionally assigns quality scores based on the entire image rather than the edited region, which may filter out some higher-quality edits. However, this does not harm the overall dataset quality, as our guiding principle is to err on the side of stricter filtering—prefer removing a few good samples rather than keeping any low-quality ones.

\section{Segmentation on different base images}
Here we provide more detailed results on the real- and AI-based subsets. As shown in~\cref{tab:supp_exp_subset}, Deeplabv3+ performs generally better on AI-based images than on real-image bases.

\input{tables/supp_exp_subset}

\section{Additional visualization of edited images}
\begin{figure*}[t]
    \centering
    \includegraphics[width=0.95\linewidth]{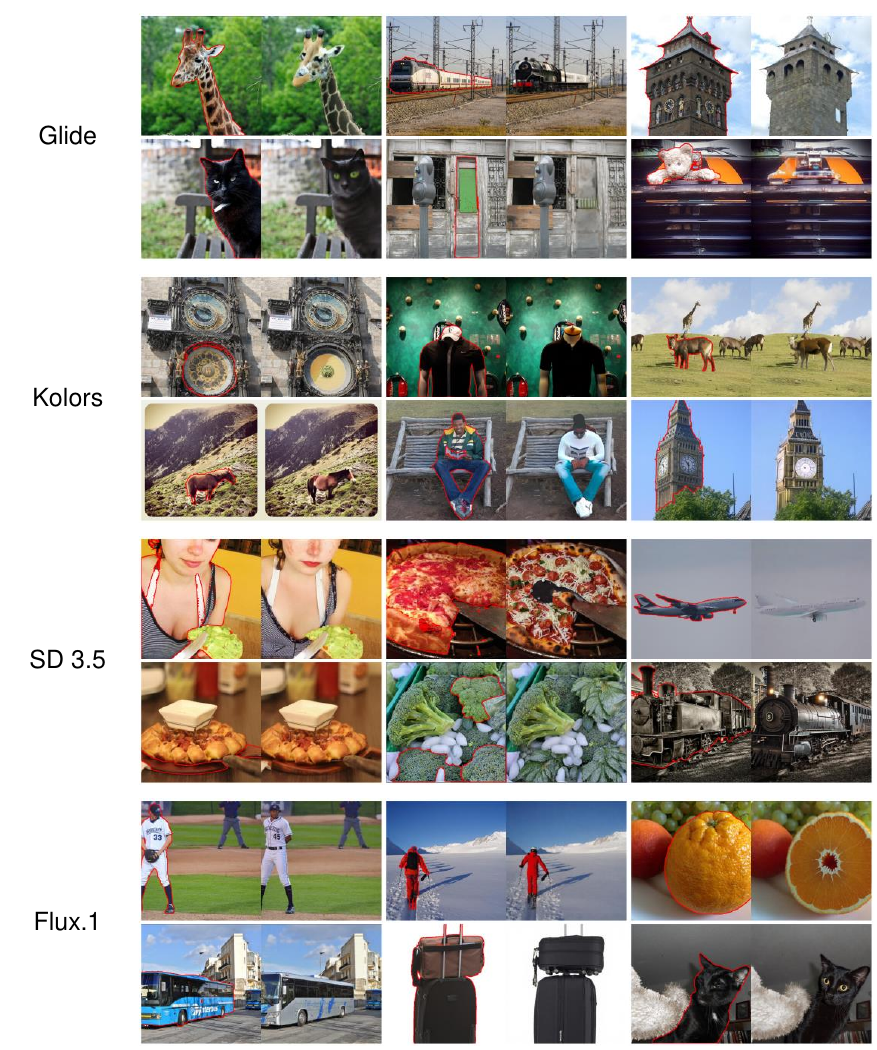}
    \caption{\textbf{More examples in DiffSeg30k.} For each pair of images, the left is the original image with red contours highlighting areas to be edited. The right is the editing results.}
    \label{fig:supp_edit_qual_1}
\end{figure*}

\begin{figure*}[t]
    \centering
    \includegraphics[width=0.95\linewidth]{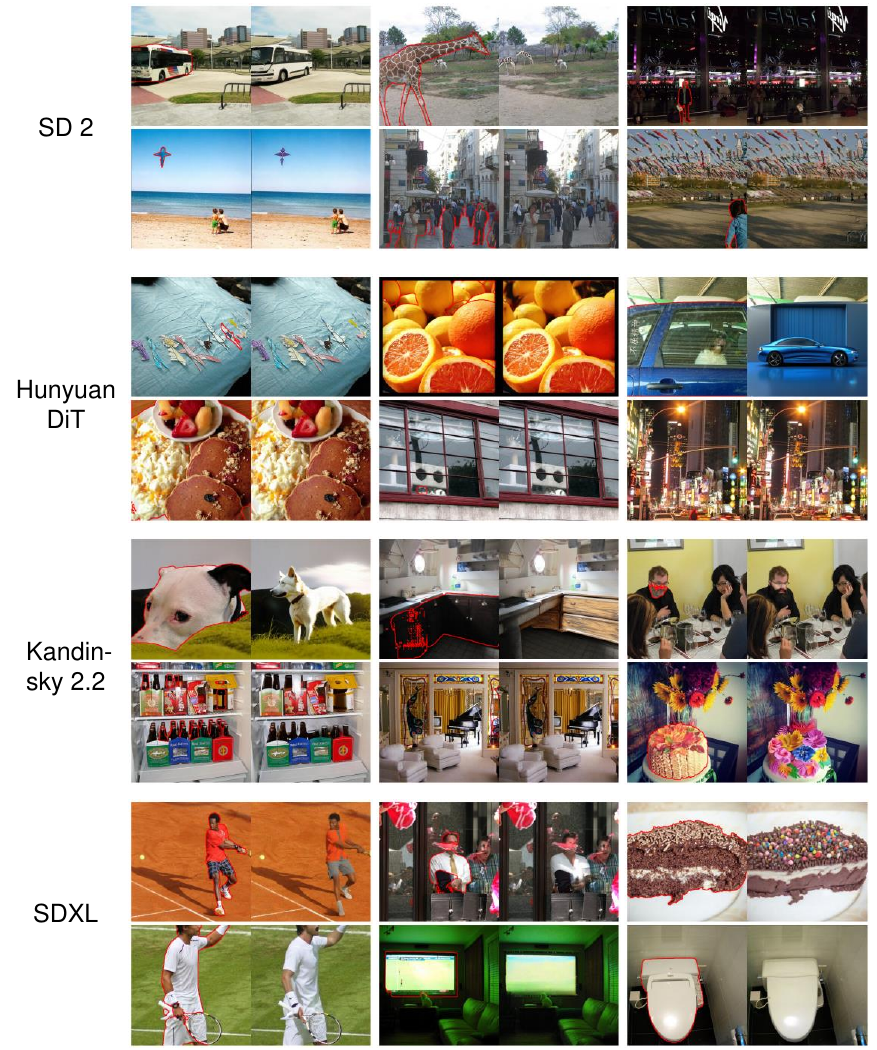}
    \caption{\textbf{More examples in DiffSeg30k.} For each pair of images, the left is the original image with red contours highlighting areas to be edited. The right is the editing results.}
    \label{fig:supp_edit_qual_2}
\end{figure*}

\cref{fig:supp_edit_qual_1,fig:supp_edit_qual_2} present additional visualizations of edited images from the DiffSeg30k dataset, showcasing the diversity of edit types and visual characteristics across different diffusion models.


%% file: tables/supp_lora.tex

\begin{table}[ht]
\centering
\caption{\textbf{Effect of LoRA on localization performance.} We evaluate semantic segmentation model Deeplabv3+, trained on SDXL-edited images and validated on SDXL- and SDXL-LoRA-edited images. We report mIoU only on masks from SDXL or SDXL-LoRA edits to isolate LoRA's impact. \textcolor{gray}{Gray} colors the baseline.}
\label{tab:supp_lora}
\begin{tabular}{lll}
\toprule
                             & Edit model (in val) & mIoU (SDXL) \\ \hline
\multirow{2}{*}{Deeplab v3+~\citep{chen2017rethinking}} & \textcolor{gray}{SDXL}       & \textcolor{gray}{0.657}       \\
                             & SDXL-LoRA  & 0.630       \\ 
                             \bottomrule
\end{tabular}
\end{table}

%% file: tables/supp_exp_subset.tex
\begin{table}
\centering
\caption{\textbf{Performance of Deeplab v3+ on real- and AI- based images.} We report binary segmentation (localizing edited regions) and semantic segmentation (localizing and attributing edits to specific diffusion models).}
\label{tab:supp_exp_subset}
\begin{adjustbox}{width=\linewidth}
\begin{tabular}{lllllll}
\toprule
& \multicolumn{3}{l}{Binary segmentation} & \multicolumn{3}{l}{Semantic segmentation} \\ \cline{2-7}
Subset  & Acc & mIoU & bF1 & Acc & mIoU & bF1 \\ 
\midrule
Real-base & 0.964 & 0.911 & 0.506 & 0.918 & 0.731 & 0.339 \\
AI-base   & 0.999 & 0.999 & 0.999 & 0.887 & 0.760 & 0.516 \\
\bottomrule
\end{tabular}
\end{adjustbox}
\end{table}
